%% file: main.tex
\documentclass[10pt,twocolumn,letterpaper]{article}

\usepackage[pagenumbers]{cvpr} 

\usepackage{graphicx}
\usepackage{amsmath}
\usepackage{amssymb}
\usepackage{siunitx}
\usepackage{hhline}
\usepackage{subcaption}
\usepackage{csquotes}
\usepackage{multirow}
\usepackage{pifont}
\usepackage[super]{nth}
\usepackage{array}
\usepackage{dcolumn}
\usepackage{booktabs}

\newcolumntype{L}[1]{>{\raggedright\let\newline\\\arraybackslash\hspace{0pt}}m{#1}}
\newcolumntype{C}[1]{>{\centering\let\newline\\\arraybackslash\hspace{0pt}}m{#1}}
\newcolumntype{R}[1]{>{\raggedleft\let\newline\\\arraybackslash\hspace{0pt}}m{#1}}

\definecolor{darkergreen}{RGB}{21, 152, 56}
\definecolor{red2}{RGB}{252, 54, 65}
\newcommand{\cmark}{\textcolor{darkergreen}{\ding{51}}}
\newcommand{\xmark}{\textcolor{red2}{\ding{55}}}

\definecolor{cvprblue}{rgb}{0.21,0.49,0.74}
\usepackage[pagebackref,breaklinks,colorlinks,allcolors=cvprblue]{hyperref}

\def\paperID{33004} 
\def\confName{CVPR}
\def\confYear{2026}

\title{Enhancing Hands in 3D Whole-Body Pose Estimation with \\Conditional Hands Modulator}

\author{Gyeongsik Moon\\
Korea University\\
{\tt\small mks0601@korea.ac.kr}\\
\small \url{https://mks0601.github.io/Hand4Whole-plus-plus}
}

\begin{document}
\maketitle

\begin{abstract}
Accurately recovering hand poses within the body context remains a major challenge in 3D whole-body pose estimation.
This difficulty arises from a fundamental supervision gap: whole-body pose estimators are trained on full-body datasets with limited hand diversity, while hand-only estimators, trained on hand-centric datasets, excel at detailed finger articulation but lack global body awareness.
To address this, we propose \textbf{Hand4Whole++}, a modular framework that leverages the strengths of both pre-trained whole-body and hand pose estimators.
We introduce CHAM (Conditional Hands Modulator), a lightweight module that modulates the whole-body feature stream using hand-specific features extracted from a pre-trained hand pose estimator.
This modulation enables the whole-body model to predict wrist orientations that are both accurate and coherent with the upper-body kinematic structure, without retraining the full-body model.
In parallel, we directly incorporate finger articulations and hand shapes predicted by the hand pose estimator, aligning them to the full-body mesh via differentiable rigid alignment.
This design allows Hand4Whole++ to combine globally consistent body reasoning with fine-grained hand detail.
Extensive experiments demonstrate that Hand4Whole++ substantially improves hand accuracy and enhances overall full-body pose quality.
\end{abstract}

\input{src/introduction}

\input{src/related_works}

\input{src/hand4whole++}

\input{src/experiments}

\input{src/conclusion}

\clearpage

\input{main_suppl}

\section*{Acknowledgments}
This work was partly supported by the Institute of Information \& Communications Technology Planning \& Evaluation(IITP)-ICT Creative Consilience Program grant funded by the Korea government(MSIT)(IITP-2026-RS-2020-II201819, 20\%).
This research was supported by the Culture, Sports and Tourism R\&D Program through the Korea Creative Content Agency grant funded by the Ministry of Culture, Sports and Tourism in 2026(Project Name: Development of AI-based image expansion and service technology for high-resolution (8K/16K) service of performance contents, Project Number: RS-2024-00395886, Contribution Rate: 20\%).
This work was supported by the Industrial Technology Innovation Program(RS-2025-02653087, Development of a Motion Data Collection System and Dynamic Persona Modeling Technology) funded By the Ministry of Trade, Industry \& Energy(MOTIE, Korea).
This work was supported by the IITP grant funded by the MSIT (No. RS-2025-25441838, Development of a human foundation model for human-centric universal artificial intelligence and training of personnel).
This work was supported by the National Research Foundation of Korea(NRF) grant funded by the Korea government(MSIT)(RS-2025-21063115).

\clearpage

{
    \small
    \bibliographystyle{ieeenat_fullname}
    \bibliography{main}
}


\end{document}

%% file: src/introduction.tex
\section{Introduction}

Accurate 3D whole-body pose estimation is crucial for a wide range of applications, including human-robot interaction, virtual avatar control, and motion understanding.
While recent approaches~\cite{choutas2020monocular,feng2021collaborative,rong2021frankmocap,moon2022accurate,li2025hybrik,zhang2023pymaf,sun2024aios,cai2023smpler,lin2023one} have made significant progress by leveraging parametric whole-body models such as SMPL-X~\cite{pavlakos2019expressive}, they still struggle to recover fine-grained hand articulations within the full-body context.

\begin{figure}[t]
\begin{center}
\includegraphics[width=\linewidth]{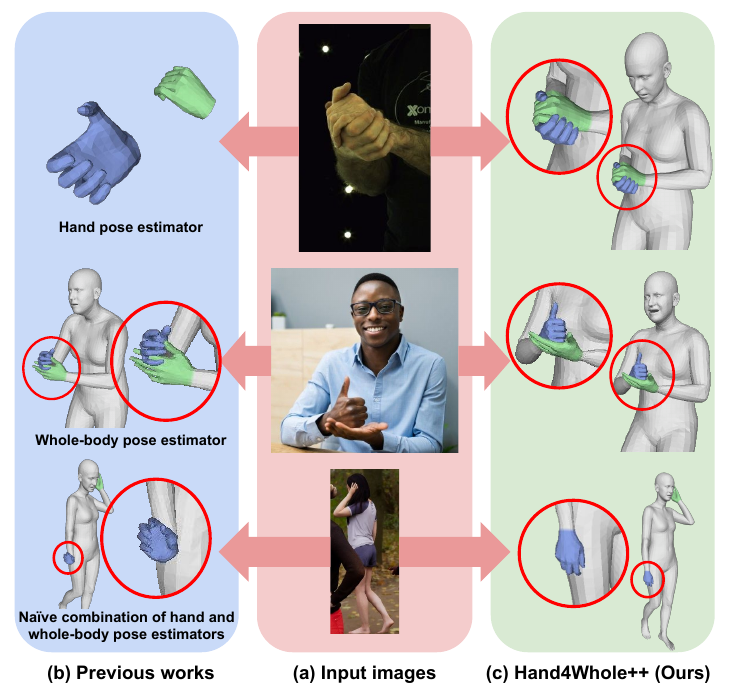}
\end{center}
\vspace*{-5mm}
\caption{
Comparison between (b) previous works and (c) the proposed Hand4Whole++.
Hand pose estimators~\cite{pavlakos2024reconstructing,potamias2024wilor} recover each hand well but fail under interaction due to missing full-body context.
Whole-body pose estimators~\cite{cai2023smpler} lack hand accuracy due to limited hand diversity in whole-body training data.
Naïvely combining both leads to implausible hands, especially under occlusion.
In contrast, Hand4Whole++ recovers accurate and plausible hands within full-body context.
}
\vspace*{-3mm}
\label{fig:intro_compare}
\end{figure}

As shown in Fig.~\ref{fig:intro_compare}, existing methods struggle to recover accurate and coherent hand poses when reasoning within the full-body context.  
Hand pose estimators~\cite{pavlakos2024reconstructing,potamias2024wilor} are effective at predicting each hand in isolation, but often fail to produce plausible results when hands interact or must be coordinated with the global body pose.  
Whole-body pose estimators~\cite{cai2023smpler} capture overall structure but lack the capacity to represent fine-grained hand articulations, frequently resulting in coarse hand predictions.
A workaround is to naïvely combine the two by attaching hand pose estimator outputs to the body~\cite{rong2021frankmocap}, which often leads to physically implausible hand placements—especially when hands are occluded or not visible.

This limitation arises from a mismatch in training supervision: whole-body pose estimators are typically trained on datasets with full-body annotations~\cite{Patel:CVPR:2021,black2023bedlam,lin2014microsoft,lin2023one,fan2023arctic}, but with limited diversity in hand poses and shapes.  
Conversely, hand-centric datasets~\cite{Freihand2019,moon2020interhand2,moon2023dataset,pavlakos2024reconstructing,fan2023arctic,kwon2021h2o,hampali2020honnotate,chao2021dexycb} provide detailed hand annotations but do not include corresponding full-body labels.  
This disconnect in supervision prevents models from learning how hands behave and interact within the full-body pose space, leading to the qualitative failures shown in the figure.

To address these challenges, we propose \textbf{Hand4Whole++}, a modular framework that integrates pre-trained whole-body and hand pose estimators to enhance hand prediction within full-body 3D pose estimation.
We introduce \textbf{CHAM (Conditional Hands Modulator)}, a lightweight module that modulates the whole-body feature stream using hand-specific features extracted from a pre-trained hand pose estimator.
This modulation enables the whole-body model to predict wrist orientations that are both accurate and consistent with the upper-body kinematic structure, without requiring retraining.
During training, both the whole-body and hand pose estimators are kept frozen, and only CHAM is optimized.
This preserves the capabilities of the pre-trained models while enabling effective adaptation to hand-centric cues without compromising generalization.

In addition, we directly incorporate finger articulations and hand shapes predicted by the hand pose estimator.
Specifically, we generate a canonical hand mesh from the MANO~\cite{romero2017embodied} parameters and rigidly align it to the wrist orientation predicted by the whole-body model using a differentiable transformation.
This enables the recovery of fine-grained finger articulations, which are difficult to achieve through full-body supervision alone.
Moreover, it improves hand shape accuracy by leveraging the specialized shape space of the hand-only model, in contrast to whole-body models~\cite{pavlakos2019expressive}, which jointly encode the body, hands, and face in a shared latent space.
This design effectively combines the global structural reasoning of the whole-body model with the precise articulation and shape expressiveness of the hand model.

Extensive experiments demonstrate that Hand4Whole++ significantly improves hand accuracy compared to state-of-the-art whole-body methods, and that this enhancement also leads to improved overall whole-body pose estimation.
Our results highlight the importance of part-aware, modular enhancement for bridging annotation gaps across heterogeneous datasets, while maintaining the efficiency and generalization capabilities of the pre-trained estimators.

Our key contributions are as follows:
\begin{itemize}
\item We propose \textbf{Hand4Whole++}, a modular framework that integrates pre-trained whole-body and hand pose estimators to improve hand accuracy within full-body 3D pose estimation.
\item We introduce \textbf{CHAM}, a lightweight feature modulation module that injects hand-specific features into a frozen whole-body pose estimator, enabling accurate wrist orientation and coherent upper-body predictions without retraining.
\item We incorporate fine-grained finger articulations and expressive hand shapes from the hand pose estimator via differentiable rigid alignment, leveraging the specialization of hand-centric models.
\end{itemize}

%% file: src/related_works.tex
\section{Related works}

\noindent\textbf{3D whole-body pose estimation.}
A common approach is to regress SMPL-X parameters, including 3D joint rotations, facial expressions, and body shape~\cite{pavlakos2019expressive}.
Several architectures have been proposed: FrankMocap~\cite{rong2021frankmocap} uses separate body and hand networks; Hand4Whole~\cite{moon2022accurate} fuses features at the joint level; OSX~\cite{lin2023one} adopts a one-stage design; PyMAF-X~\cite{zhang2023pymaf}, HybrIK-X~\cite{li2025hybrik}, and AiOS~\cite{sun2024aios} introduce various model-specific improvements; and SMPLer-X~\cite{cai2023smpler} is trained as a foundation model with 4.5M poses.
Despite this progress, hand accuracy remains limited compared to hand-specific models.
DOPE~\cite{weinzaepfel2020dope} distills part-specific experts but cannot leverage hand-only datasets~\cite{moon2020interhand2,Freihand2019,moon2023dataset} due to missing full-body context.
HMR-Adapter~\cite{shen2024hmr} addresses a similar problem but struggles with interacting hands due to its neglect of relative hand positions. While HMR-Adapter refines the body using hand features interpolated from internal whole-body features—which are often uninformative in hand-centric images—our Hand4Whole++ bridges this gap. Specifically, we inject informative features from an external hand model into the whole-body stream through CHAM, incorporating hand-specific precision into full-body estimation.

\begin{figure*}[t]
\begin{center}
\includegraphics[width=\linewidth]{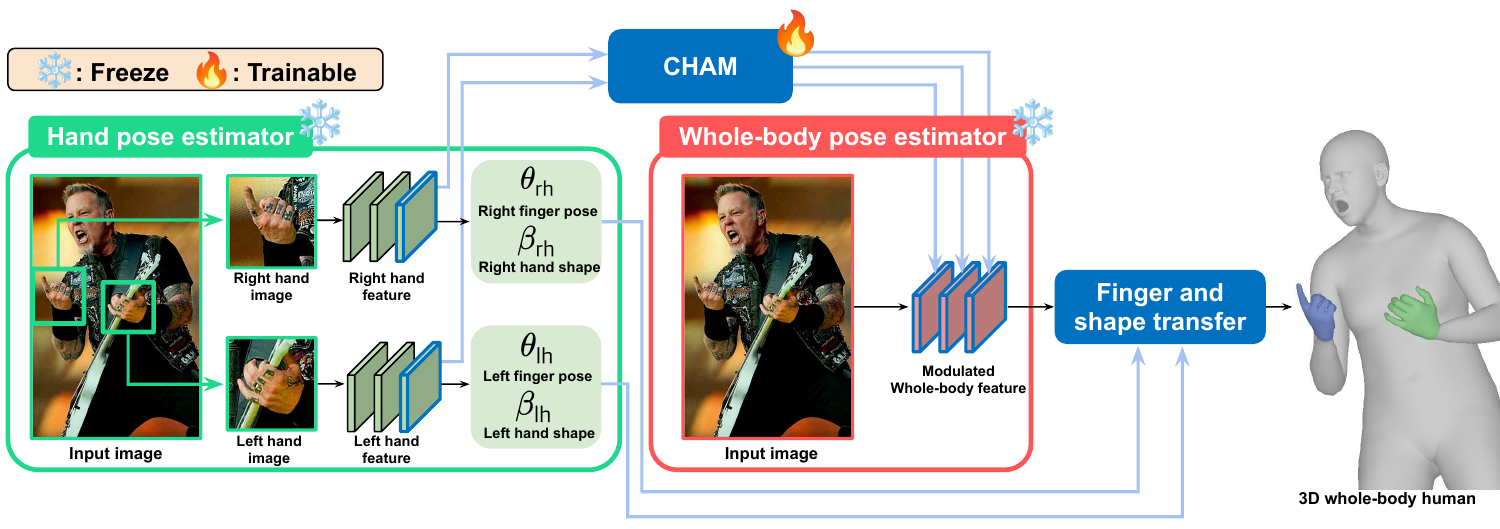}
\end{center}
\vspace*{-5mm}
\caption{
Overview of Hand4Whole++, which comprises a pre-trained hand pose estimator, a pre-trained whole-body pose estimator, CHAM, and a finger articulation and shape transfer module.
During training, only CHAM is updated, while the pre-trained pose estimators remain frozen.
}
\vspace*{-3mm}
\label{fig:overview}
\end{figure*}

\noindent\textbf{3D hand pose estimation.}
A standard approach is to regress MANO parameters~\cite{romero2017embodied}, including 3D finger rotations and low-dimensional shape.
Recent works explore various representations and architectures to improve accuracy.
I2L-MeshNet~\cite{moon2020i2l} uses a lixel-based representation to predict 3D mesh vertices; Pose2Mesh~\cite{choi2020p2m} lifts 2D keypoints to 3D using graph convolutions.
Transformer-based methods like METRO~\cite{lin2021end} and HaMeR~\cite{pavlakos2024reconstructing} enable global context reasoning and improve generalization.
WiLoR~\cite{potamias2024wilor} focuses on refinement of initial predictions.
These models show strong performance on challenging benchmarks~\cite{moon2020interhand2,chao2021dexycb,fan2023arctic}, highlighting the value of hand-specific supervision and design.

\noindent\textbf{Conditional modulation.}
Controlling pre-trained models with lightweight modules has become a popular strategy for improving versatility without full retraining.
These methods inject task-specific signals into frozen backbones, reducing computation.
Examples include ControlNet~\cite{zhang2023adding}, T2I-Adapter~\cite{mou2024t2i}, and X-Adapter~\cite{ran2024x} for image models, and ControlVideo~\cite{zhang2023controlvideo}, CTRL-Adapter~\cite{lin2024ctrl} for video.
Inspired by this, CHAM modulates whole-body features using hand-specific cues while keeping both pose estimators frozen.

%% file: src/hand4whole++.tex
\section{Hand4Whole++}

\subsection{Overview}

Fig.~\ref{fig:overview} illustrates the overall architecture of the proposed Hand4Whole++.
Our goal is to improve 3D hand accuracy within a whole-body pose estimation pipeline by leveraging the strengths of hand-only pose estimators.
Hand4Whole++ is a modular framework that integrates a pre-trained whole-body pose estimator and a pre-trained hand pose estimator, enhanced by a lightweight modulation module called CHAM.
In our implementation, we adopt SMPLer-X-L32~\cite{cai2023smpler} as the whole-body pose estimator and WiLoR~\cite{potamias2024wilor} as the hand pose estimator, both chosen for their strong performance on their respective tasks.

The framework operates in a fully feed-forward manner.
Given an input image, the hand pose estimator extracts hand features and predicts hand pose and shape parameters.
CHAM then modulates the whole-body image feature stream using the extracted hand features as conditional signals.
From the modulated features, the whole-body pose estimator predicts the full set of SMPL-X parameters, including body, hand, and face pose, as well as shape.

To obtain high-fidelity hand predictions, we discard the finger pose parameters predicted by the whole-body estimator.
Instead, we use the detailed hand mesh predicted by the hand pose estimator, which is defined in a canonical wrist orientation space.
This mesh is rigidly aligned to the wrist orientation predicted by the whole-body model, allowing us to incorporate the more accurate finger articulations and expressive hand shape from the hand estimator while preserving consistency with the global body pose.

For jaw pose and facial expression regression, we use the pre-trained FaceNet from Hand4Whole~\cite{moon2022accurate}, which takes a cropped face image as input based on facial keypoints from the whole-body pose estimator.
FaceNet, based on ResNet-18~\cite{he2016deep}, is lightweight and achieves better performance than the face branch of the whole-body estimator given the limited facial diversity in CHAM's training set.

\subsection{Conditional hands modulator (CHAM)}\label{subsec:cham}

\begin{figure}[t]
\begin{center}
\includegraphics[width=\linewidth]{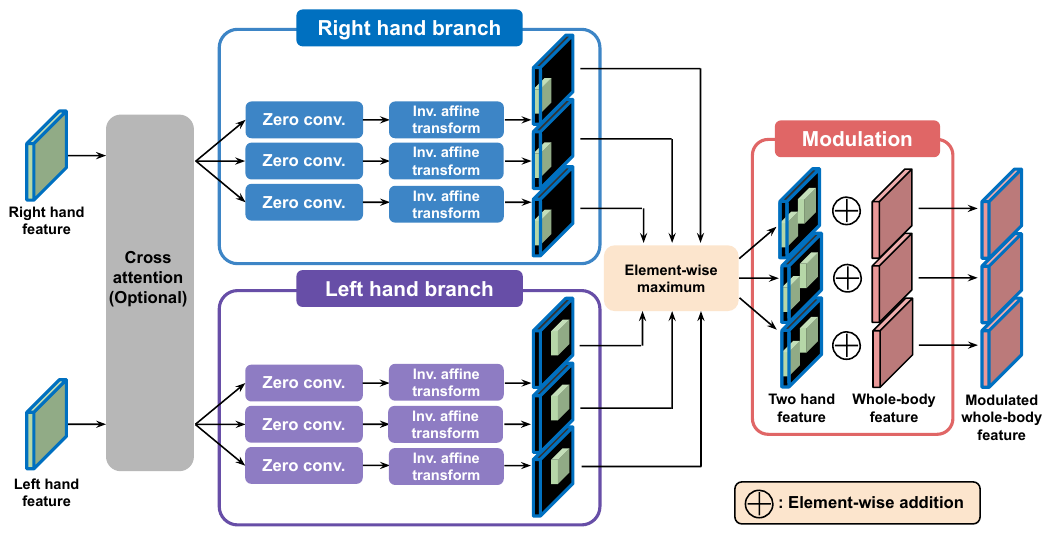}
\end{center}
\vspace*{-5mm}
\caption{
Architecture of the proposed CHAM. The gray dashed box (2D positional encoding and cross-attention) is used only when both hands are detected. 
Otherwise, each hand feature is directly passed to its corresponding branch. 
For simplicity, we illustrate only three layers instead of the full 24-layer design. 
}
\vspace*{-3mm}
\label{fig:cham}
\end{figure}

Fig.~\ref{fig:cham} illustrates the architecture of CHAM, a lightweight module that modulates whole-body features using hand-specific information in a spatially aligned manner.
Inspired by ControlNet~\cite{zhang2023adding}, CHAM enables conditional modulation of a frozen pre-trained model, improving wrist orientation prediction within the upper-body kinematic chain without disrupting the whole-body pose estimation capability learned from large-scale datasets.
While we use SMPLer-X~\cite{cai2023smpler} and WiLoR~\cite{potamias2024wilor} in our implementation, CHAM is compatible with other ViT-based whole-body models, as it operates on transformer tokens without architecture-specific dependencies.
CHAM adds only 10 ms, about 10\% of the total runtime at 10 frames per second.

\noindent\textbf{Architecture.}
Given an input image, we first localize the hand regions~\cite{yang2023effective,potamias2024wilor} and crop them to a resolution of $256 \times 256$, as required by WiLoR.
These cropped images are passed through the pre-trained WiLoR hand pose estimator, which uses a ViT backbone~\cite{dosovitskiy2021image} to predict hand pose and shape parameters.
We extract the final-layer ViT features for both hands (the leftmost right and left hand features in Fig.~\ref{fig:cham}) and use them as conditional inputs to CHAM as it contains essential hand articulation information.
For non-detected hands, we set the ViT features to zero.

CHAM processes the hand features by first applying 2D positional encodings, followed by a three-layer cross-attention Transformer encoder~\cite{vaswani2017attention}.
The 2D positional encoding is generated from the original full-body image space and then crop-and-resize is applied using the same bounding box used for hand image preparation, preserving the spatial information of the hands within the global body context.
Cross-attention is applied only when both hands are detected, allowing CHAM to model inter-hand relationships.
If only one hand, or neither hand, is detected, CHAM bypasses both the 2D positional encoding and cross-attention, and directly uses the available ViT features.

The resulting hand features are passed through two separate branches in CHAM: a left-hand branch and a right-hand branch.
Each branch contains a set of 24 independent $1\times1$ convolutional layers, corresponding to the 24 Transformer blocks in SMPLer-X.
Specifically, the left-hand features are processed only by the convolutional layers in the left-hand branch, and the right-hand features by those in the right-hand branch.
Each convolutional layer processes its input in parallel, producing a distinct hand feature for its corresponding transformer block.
All convolution layers are initialized to zero, following the design of ControlNet~\cite{zhang2023adding}, ensuring that CHAM starts from a neutral state and learns to modulate only when beneficial.

To maintain spatial alignment with the whole-body feature map, CHAM applies an inverse affine transformation to each hand feature, undoing the crop-and-resize operation used during hand localization for WiLoR~\cite{potamias2024wilor}.
Zero padding is applied to fill non-hand regions.
The aligned outputs from the left and right branches are then merged using an element-wise maximum operation to form two-hand features.
CHAM produces a list of 24 two-hand features, which are additively fused into the spatial tokens of the ViT feature stream at each corresponding block of SMPLer-X to modulate the whole-body representation.

During training, both the SMPLer-X and WiLoR pose estimators remain frozen to preserve their pre-trained capabilities.
Only CHAM is optimized, enabling efficient and targeted learning without requiring joint training or full-body annotations in hand-centric datasets.

\noindent\textbf{Modulating upper-body kinematics for hand orientation.}
The primary role of CHAM is to improve wrist orientation within the full-body context, rather than detailed finger articulation, which is addressed separately in the next subsection.
A naïve baseline for combining whole-body and hand-only pose estimators to obtain wrist orientation would be to directly assign the wrist orientation (\emph{i.e.}, wrist orientation in the camera coordinate system) predicted by the hand pose estimator to the SMPL-X wrist joint.
However, this approach often results in anatomically implausible wrist configurations, as the wrist orientation from the hand pose estimator is estimated independently of the body and lacks awareness of the upper-body kinematic chain.
In practice, this leads to disjointed or physically invalid wrist orientations that are inconsistent with the shoulder and elbow joints.
CHAM addresses this issue by modulating the whole-body features with hand-specific cues, enabling the body model to predict wrist poses that are both accurate and anatomically coherent with the global body structure.
This targeted modulation allows seamless integration of hand-specific information into the full-body estimation pipeline, without compromising structural plausibility.
Conversely, injecting body features into the hand stream only affects the wrists, leaving other body joints unmodulated. 
Such restricted modulation fails to refine relative hand positioning, leading to suboptimal performance.

\begin{figure}[t]
\begin{center}
\includegraphics[width=\linewidth]{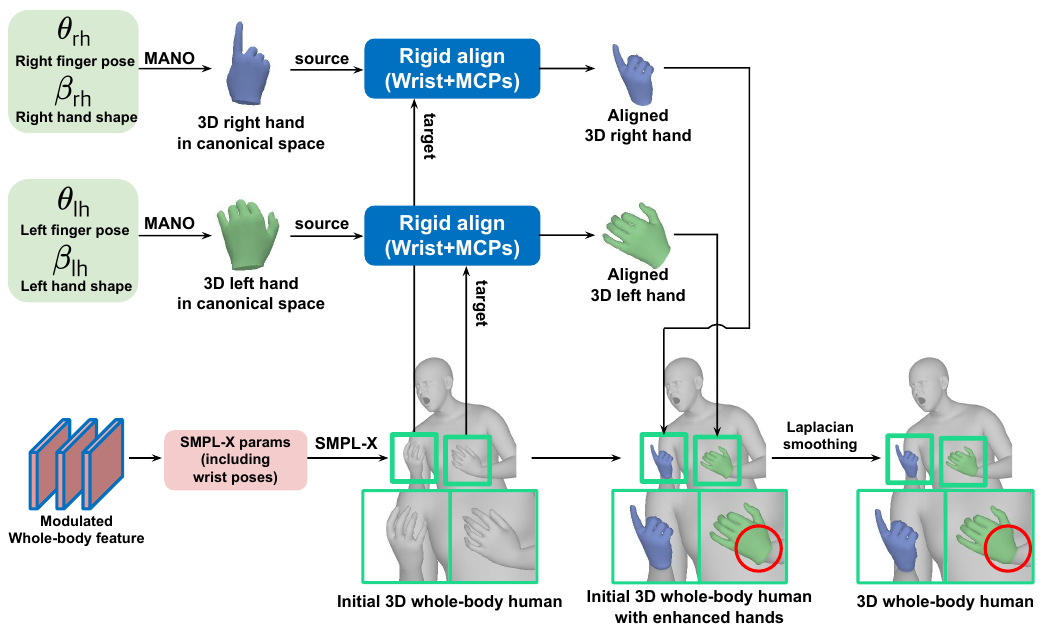}
\end{center}
\vspace*{-5mm}
\caption{
Pipeline of the finger and shape transfer. 
We align the canonical 3D hand mesh to the initial whole-body mesh using the wrist and the four MCP joints (index, middle, ring, and pinky).
}
\vspace*{-3mm}
\label{fig:finger_shape_transfer}
\end{figure}

\subsection{Finger articulation and hand shape transfer}\label{subsec:finger_shape_transfer}

While CHAM modulates the whole-body feature stream to enable accurate and anatomically coherent wrist orientation, it does not model finger articulation or hand shape.
Instead, as shown in Fig.~\ref{fig:finger_shape_transfer}, we directly transfer these components from the hand pose estimator, which is trained on hand-centric datasets and specializes in predicting accurate finger articulation and hand shapes.

We use the MANO parameters predicted by the hand pose estimator—specifically, finger pose ($\theta_\text{rh}$ and $\theta_\text{lh}$) and shape parameters ($\beta_\text{rh}$ and $\beta_\text{lh}$)—while ignoring the wrist orientation.
Using these parameters, we generate a hand mesh and keypoints in the canonical (zero) wrist orientation space.
To align this hand mesh with the SMPL-X full-body mesh, we perform a rigid alignment based on the wrist and four MCP joints of the index, middle, ring, and pinky fingers.
These joints provide a stable and anatomically meaningful reference for aligning the canonical hand to the wrist region of the SMPL-X mesh~\cite{moon2020deephandmesh,moon2024authentic}.
The resulting rigid transformation is applied to the canonical MANO hand mesh, which then replaces the corresponding hand vertices of the SMPL-X mesh.

Importantly, this alignment step is fully differentiable, allowing gradients to flow back through CHAM for the wrist orientation estimation.
Although the hand pose estimator also predicts a wrist orientation, we discard it entirely to avoid inconsistency with the full-body kinematic chain.
Instead, wrist position and orientation are determined solely by the SMPL-X model, which is modulated by CHAM.
This design enables precise articulation and shape from the hand estimator to be combined with anatomically coherent hand placement from the body estimator, while ensuring spatial and structural consistency through differentiable alignment.

To further improve geometric continuity between the inserted MANO hand mesh and the surrounding SMPL-X wrist region, we apply a simple Laplacian smoothing to the hand boundary vertices after alignment.
This helps mitigate potential seam artifacts caused by differences in wrist geometry between the two models, resulting in a smoother and more visually consistent integration.

\subsection{Loss Functions}

During training, we freeze both the SMPL-X and hand pose estimators and optimize only CHAM.
The objective is to encourage anatomically consistent wrist orientation within the full-body context by supervising the upper-body kinematic chain through a combination of pose and keypoint losses.

\noindent\textbf{Pose loss.}
For full-body datasets that provide SMPL-X pose annotations, we minimize the $\ell_1$ distance between the estimated and ground-truth (GT) 3D joint rotations.
For hand-only datasets that provide wrist orientation annotations, we convert the local wrist pose in SMPL-X to global wrist orientation via forward kinematics and supervise them accordingly.

\noindent\textbf{Shape loss.}
For full-body datasets with SMPL-X shape annotations, we minimize the $\ell_1$ distance between the estimated and GT shape parameters.
For hand-only datasets, we apply an $\ell_2$ regularization penalty to the SMPL-X shape parameters to prevent unrealistic shape predictions.

\noindent\textbf{2D and 3D keypoint losses.}
We compute $\ell_1$ losses on both 2D and 3D keypoints.
For the 2D loss, 3D keypoints are projected into image space and supervised against GT 2D coordinates.
For 3D keypoints, losses are computed in different reference frames depending on the dataset: (1) pelvis-relative for full-body datasets, (2) right-wrist-relative for interacting-hand datasets, and (3) wrist-relative for hand-only datasets.

\noindent\textbf{Body root pose regularization.}
For hand-only samples that lack full-body annotations, we regularize the SMPL-X root pose to maintain a vertical torso orientation in the world coordinate system.
This mitigates ambiguity caused by cropped hand images and helps stabilize upper-body predictions.

%% file: src/experiments.tex
\section{Experiments}

\subsection{Protocol}~\label{subsec:protocol}

\noindent\textbf{Datasets.}
We train Hand4Whole++ using InterHand2.6M (IH26M)~\cite{moon2020interhand2}, ReInterHand (ReIH)~\cite{moon2023dataset}, ARCTIC~\cite{fan2023arctic}, and AGORA~\cite{Patel:CVPR:2021}.
We evaluate our model on the test splits of these datasets.
Additionally, we evaluate on EHF~\cite{pavlakos2019expressive} and HIC~\cite{tzionas2016capturing} to validate generalization capability to totally unseen domains of images.
IH26M, ReIH, and HIC provide diverse 3D hand annotations without full-body annotations.
AGORA, ARCTIC, and EHF provides diverse whole-body annotations.

\noindent\textbf{Evaluation metrics.}
Following prior works~\cite{moon2022accurate,li2025hybrik,zhang2023pymaf,cai2023smpler,moon2020interhand2,moon2023dataset}, we report the mean per-vertex position error (MPVPE) and mean relative-root position error (MRRPE), both measured in millimeters.
For full-body evaluation, MPVPE is computed after aligning the pelvis (root joint).
For hand evaluation, MPVPE is computed after aligning the wrist, and MRRPE is computed between right and left wrist positions to measure relative position between two hands.
To compute MPVPE and MRRPE for hands, we extract hand vertices from the recovered SMPL-X mesh using the official hand vertex indices provided by the SMPL-X model.
We do not report PA-MPVPE (Procrustes-Aligned MPVPE) for hands, as it removes global rotation errors, thereby ignoring wrist orientation—which is one of the primary goals of our CHAM.

\begin{table*}[t]
\footnotesize
\centering
\setlength\tabcolsep{1.0pt}
\def\arraystretch{1.1}
\caption{
Comparisons of baselines and our Hand4Whole++ on full-body and hand-only datasets.
}
\vspace*{-3mm}
\scalebox{1.0}{
\begin{tabular}{L{5.5cm}|C{1.9cm}C{1.6cm}C{1.9cm}|C{2.5cm}C{2.2cm}C{1.9cm}}
\specialrule{.1em}{.05em}{.05em}
\multirow{2}{*}{Settings} & \multicolumn{3}{c|}{Full-body datasets (Full/hands MPVPE)} & \multicolumn{3}{c}{Hand-only datasets (MPVPE/MRRPE)} \\
 & AGORA & ARCTIC & EHF & IH26M & ReIH & HIC \\ \hline
Original whole-body~\cite{cai2023smpler} & 85.61/52.31 &  56.06/31.48 & 63.26/46.21 &  \phantom{0}38.64/119.56 & \phantom{0}58.86/101.82 & 32.43/67.68 \\ 
Fine-tuned whole-body~\cite{cai2023smpler} & 90.77/55.91 & 67.52/29.03 & 126.34/57.35\phantom{0} & 20.00/47.89 & 24.87/28.32 & 32.30/62.11 \\ 
Hand-only~\cite{potamias2024wilor} & \phantom{00}-\phantom{00}/99.11 & \phantom{00}-\phantom{00}/46.79 & \phantom{00}-\phantom{00}/46.28 & \phantom{000}11.17/94817.41 & \phantom{000}8.09/3094.65 & \phantom{0}\textbf{15.44}/848.23 \\
\textbf{Hand4Whole++ (Ours)} & \textbf{76.84}/\textbf{49.71} &  \textbf{45.95}/\textbf{25.03} & \textbf{61.24}/\textbf{33.43} & \phantom{0}\textbf{9.40}/\textbf{32.30} & \phantom{0}\textbf{7.98}/\textbf{16.37} & 17.72/\textbf{29.09} \\ 
\specialrule{.1em}{.05em}{.05em}
\end{tabular}
}
\vspace*{-3mm}
\label{table:ablation_fine_tune}
\end{table*}

\begin{table}[t]
\footnotesize
\centering
\setlength\tabcolsep{1.0pt}
\def\arraystretch{1.1}
\caption{MPVPE comparison on the AGORA dataset using different strategies for combining whole-body and hand-only pose estimators. The last row shows ours.}
\vspace*{-3mm}
\scalebox{1.0}{
\begin{tabular}{C{1.5cm}C{1.1cm}|C{3.0cm}C{2.0cm}}
\specialrule{.1em}{.05em}{.05em}
\multirow{2}{*}{Wrist copy} & \multirow{2}{*}{CHAM} & Full-body errors & Hands errors \\
 & & (body root-relative) & (wrist-relative) \\ \hline
\xmark & \xmark & 84.76 & 52.31 \\ 
\cmark & \xmark & 90.70 & 100.59 \\ 
\xmark & \cmark & \textbf{76.88} & \textbf{50.56} \\ 
 \specialrule{.1em}{.05em}{.05em}
\end{tabular} 
}
\vspace*{-3mm}
\label{table:ablation_whole_hand_combine}
\end{table}

\begin{figure}[t]
\begin{center}
\includegraphics[width=\linewidth]{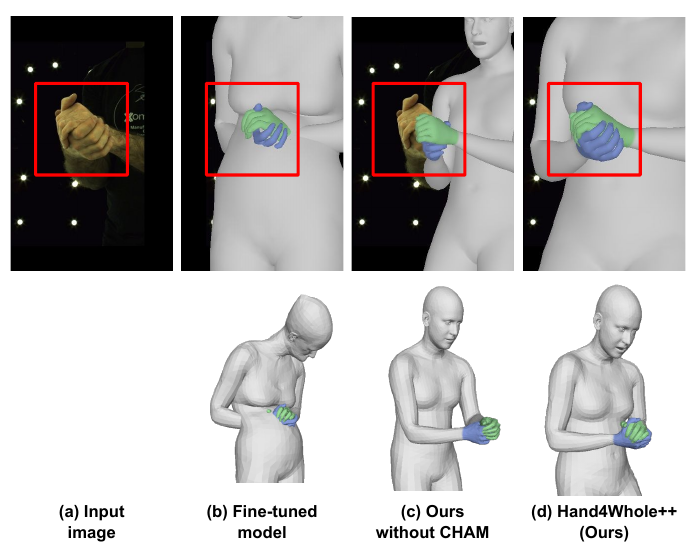}
\end{center}
\vspace*{-5mm}
\caption{
Effectiveness of the proposed CHAM.
}
\vspace*{-3mm}
\label{fig:ablation_cham}
\end{figure}

\begin{figure*}[t]
\begin{center}
\includegraphics[width=0.92\linewidth]{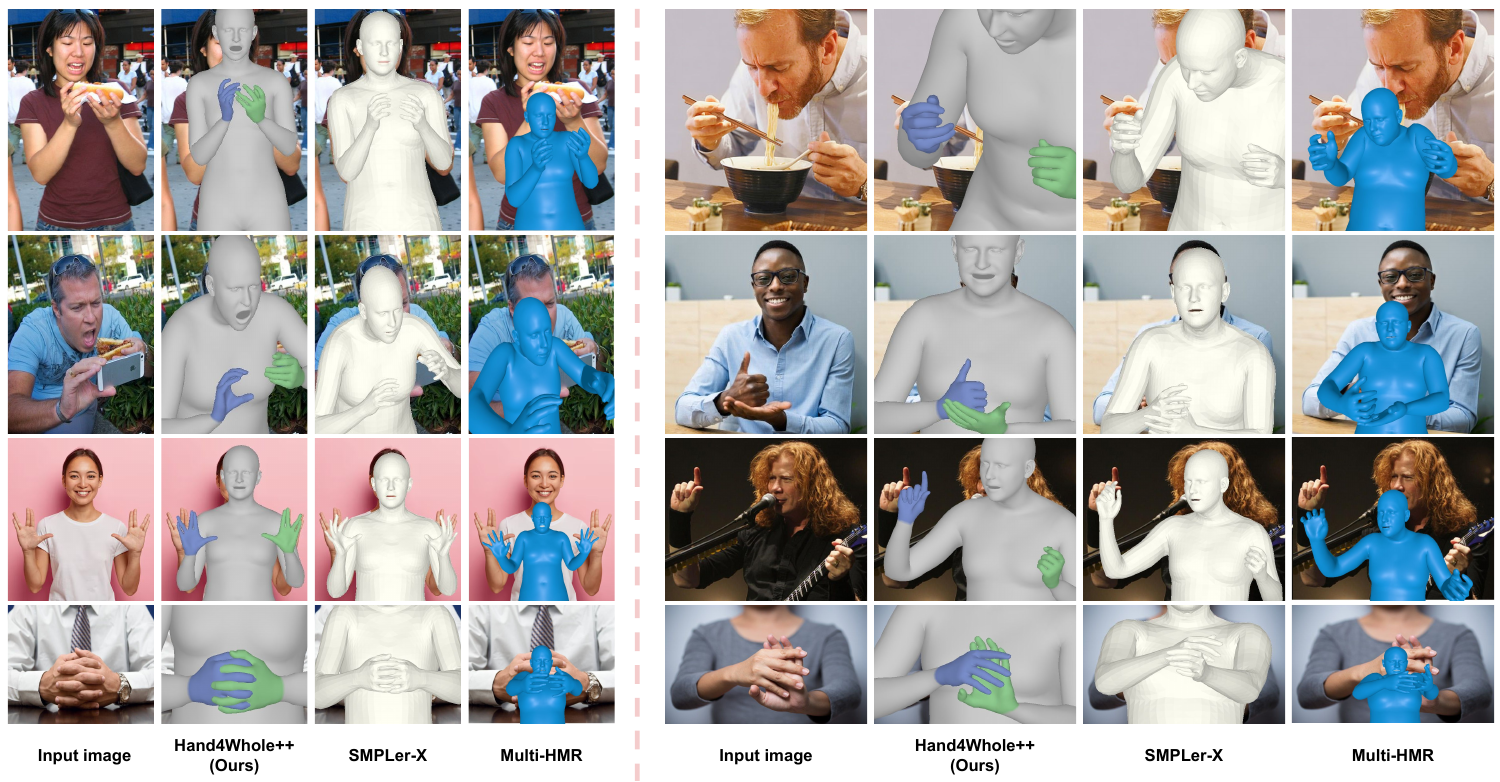}
\end{center}
\vspace*{-5mm}
\caption{
Qualitative comparison on in-the-wild images comparing our Hand4Whole++ with SMPLer-X~\cite{cai2023smpler} and Multi-HMR~\cite{baradel2024multi}.
}
\vspace*{-5mm}
\label{fig:qualitative_comparison}
\end{figure*}

\begin{figure*}[t]
\begin{center}
\includegraphics[width=0.92\linewidth]{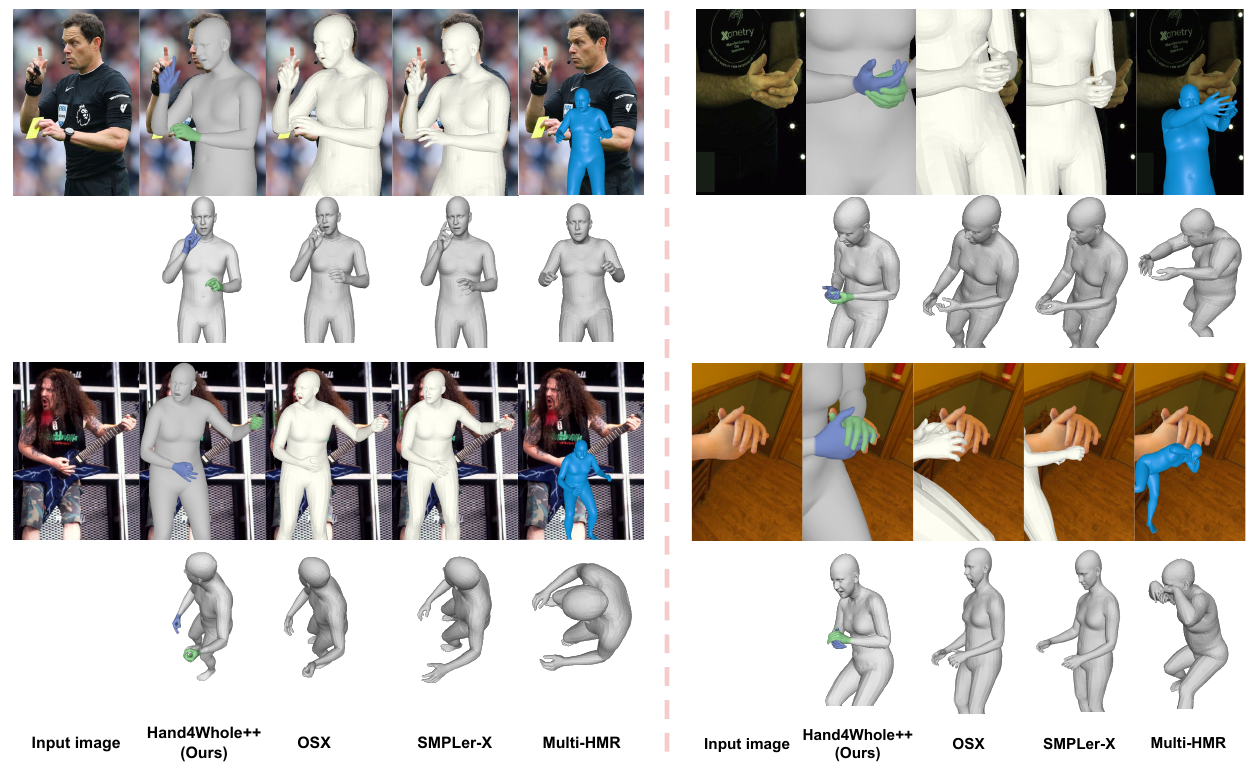}
\end{center}
\vspace*{-5mm}
\caption{
Qualitative comparison of our Hand4Whole++ with OSX~\cite{lin2023one}, SMPLer-X~\cite{cai2023smpler}, and Multi-HMR~\cite{baradel2024multi}.
}
\vspace*{-5mm}
\label{fig:qualitative_comparison_2}
\end{figure*}

\subsection{Ablation studies}

\noindent\textbf{Fine-tuning vs. CHAM.}
Tab.~\ref{table:ablation_fine_tune} shows that Hand4Whole++ achieves the lowest errors across both full-body datasets~\cite{Patel:CVPR:2021,fan2023arctic,pavlakos2019expressive} and two-hand datasets~\cite{moon2020interhand2,moon2023dataset,tzionas2016capturing}, outperforming three baselines: the original whole-body pose estimator~\cite{cai2023smpler}, its fine-tuned variant, and a state-of-the-art hand-only pose estimator~\cite{potamias2024wilor}.
The fine-tuned whole-body model is trained on the same limited dataset used for CHAM training, ensuring a fair comparison.
However, it overfits to the training data—mostly hand-only datasets—and performs poorly on unseen full-body images such as EHF.
As shown in Fig.~\ref{fig:ablation_cham} (b), it produces distorted body poses despite accurate hand alignment.
In contrast, Hand4Whole++ maintains the generalization ability of the original model while improving hand accuracy, as shown in Fig.~\ref{fig:ablation_cham} (c) and (d).
This is enabled by CHAM, which integrates hand-specific cues without modifying the pre-trained backbone.
These results demonstrate that CHAM enables efficient learning under limited supervision, enhancing hand performance while preserving global structure.
The hand-only estimator shows high accuracy in finger articulation and shape, but lacks global consistency (Fig.~\ref{fig:intro_compare}, top row).
This is reflected in its large MRRPE, indicating poor estimation of the relative positions between hands and lack of full-body awareness.
By combining the strengths of both models, Hand4Whole++ delivers accurate hand predictions while preserving overall coherence, as evidenced by its low MPVPE and MRRPE.

\noindent\textbf{How to combine whole-body and hand-only models.}
Tab.~\ref{table:ablation_whole_hand_combine} compares various strategies for combining whole-body and hand-only pose estimators.
The first row follows the strategy of Hand4Whole~\cite{moon2022accurate}, where only the finger pose is copied from the hand-only model while the wrist poses are from the pre-trained whole-body model as-is.
The second row, similar to FrankMocap~\cite{rong2021frankmocap}, directly copies the wrist orientations and finger poses from the hand-only model.
This naïve approach leads to increased error, as shown in Fig.~\ref{fig:intro_compare} bottom, because the wrist orientation predicted by the hand-only model is unaware of the full-body context, often resulting in anatomically implausible wrist configurations.

Our method in the last row combines CHAM-based modulation of the whole-body feature stream, achieving the lowest error among all settings.
Notably, while CHAM clearly reduces hand error, it also leads to a substantial reduction in full-body error.
This is because CHAM improves wrist orientation in a globally consistent manner and, more importantly, enhances the entire upper-body kinematic chain—including the shoulder, elbow, and wrist—through its influence on the whole-body feature stream (Fig.~\ref{fig:ablation_cham} (b) and (c)).
By enforcing anatomically plausible hand positioning within the body context, CHAM implicitly corrects upstream joints, contributing to better overall pose predictions.

Note that the numbers in this table differ slightly from those in Tab.~\ref{table:ablation_fine_tune}, as we disable shape transfer for all settings here to ensure fair comparison.
This omission has minimal effect on AGORA, since its SMPL-X GT offers limited diversity in hand shapes, reducing the impact of shape transfer.

\begin{table}[t]
\footnotesize
\centering
\setlength\tabcolsep{1.0pt}
\def\arraystretch{1.1}
\caption{MPVPE comparison with and without the proposed finger articulation and hand shape transfer. The last row shows ours.}
\vspace*{-3mm}
\scalebox{1.0}{
\begin{tabular}{C{1.0cm}C{1.0cm}|C{1.3cm}C{1.3cm}C{1.3cm}}
\specialrule{.1em}{.05em}{.05em}
Finger & Shape &  IH26M & ReIH & HIC \\ \hline
\xmark & \xmark & 14.69  &  18.13 & 21.68 \\ 
\cmark & \xmark &  12.26 & 15.24 & 19.61 \\ 
\cmark & \cmark & \textbf{9.40} & \textbf{7.98} & \textbf{17.72} \\
 \specialrule{.1em}{.05em}{.05em}
\end{tabular}
}
\vspace*{-3mm}
\label{table:ablation_transfer}
\end{table}

\begin{table*}[t]
\footnotesize
\centering
\setlength\tabcolsep{1.0pt}
\def\arraystretch{1.1}
\caption{
Comparisons with full-body pose estimators on full-body and hand-only datasets.
* represents evaluated with GT scale of hands following previous protocols~\cite{li2022interacting}.
}
\vspace*{-3mm}
\scalebox{1.0}{
\begin{tabular}{L{6.0cm}|C{2.0cm}C{2.0cm}C{2.0cm}|C{2.0cm}C{2.0cm}C{2.0cm}}
\specialrule{.1em}{.05em}{.05em}
\multirow{2}{*}{Methods} & \multicolumn{3}{c|}{Full-body datasets (Full/hands MPVPE)} & \multicolumn{3}{c}{Hand-only datasets (MPVPE/MRRPE)} \\
 & AGORA & ARCTIC & EHF & IH26M & ReIH & HIC \\ \hline
Hand4Whole~\cite{moon2022accurate} & 185.18/74.55\phantom{0} & 151.47/47.79\phantom{0} & 76.84/39.82 & 30.65*/219.26 & \phantom{0}71.57/310.69 & \phantom{0}22.73/101.72 \\ 
OSX~\cite{lin2023one} & 178.28/76.37\phantom{0} & 111.42/50.70\phantom{0} & 70.82/53.73 & 38.47*/173.56 & \phantom{0}71.10/221.83 & 35.51/94.57 \\ 
SMPLer-X~\cite{cai2023smpler} & 85.61/52.31 & 56.06/31.48 & 63.26/46.21 & 38.64*/119.56 & \phantom{0}58.86/101.82 & 32.43/67.68 \\
\textbf{Hand4Whole++ (Ours)} & \textbf{76.84}/\textbf{49.71} & \textbf{45.95}/\textbf{25.03} & \textbf{61.24}/\textbf{33.43} & \textbf{9.40}*/\textbf{32.30} & \phantom{0}\textbf{7.98}/\textbf{16.37} & \textbf{17.72}/\textbf{29.09} \\
\specialrule{.1em}{.05em}{.05em}
\end{tabular}
}
\vspace*{-3mm}
\label{table:compare_sota_wb}
\end{table*}

\begin{table*}[t]
\footnotesize
\centering
\setlength\tabcolsep{1.0pt}
\def\arraystretch{1.1}
\caption{
MPVPE/MRRPE comparisons with hand pose estimators on hand-only datasets.
* represents evaluated with GT scale of hands following previous protocols~\cite{li2022interacting}.
}
\vspace*{-3mm}
\scalebox{1.0}{
\begin{tabular}{L{4.5cm}|C{3.0cm}C{3.0cm}C{3.0cm}}
\specialrule{.1em}{.05em}{.05em}
Methods & IH26M & ReIH & HIC  \\ \hline
IntagHand~\cite{li2022interacting} & \textbf{9.03}*/48.04 & 19.31*/33.37\phantom{0} & 20.08*/52.46\phantom{0} \\
InterWild~\cite{moon2023bringing} & 10.28*/44.75\phantom{0} & 13.99/22.38 & 15.68/31.35 \\
HaMeR~\cite{pavlakos2024reconstructing} & \phantom{0}9.53*/594.84 & \phantom{0}17.29/644.51 & \phantom{0}16.15/526.94 \\
WiLoR~\cite{potamias2024wilor} & \phantom{00}11.17*/94817.41 & \phantom{000}8.09/3094.65 & \phantom{0}\textbf{15.44}/848.23 \\
\textbf{Hand4Whole++ (Ours)} & 9.40*/\textbf{32.30} & \phantom{0}\textbf{7.98}/\textbf{16.37} & 17.72/\textbf{29.09} \\
\specialrule{.1em}{.05em}{.05em}
\end{tabular}
}
\vspace*{-3mm}
\label{table:compare_sota_hand}
\end{table*}

\noindent\textbf{Finger articulation and hand shape transfer.}
Tab.~\ref{table:ablation_transfer} demonstrates the effectiveness of our proposed finger and shape transfer for accurate 3D hand reconstruction.
In all settings, only CHAM is trained, while the pre-trained whole-body and hand pose estimators remain frozen.
The first row uses the finger poses predicted by the whole-body model, with its features modulated by CHAM.
In this setting, CHAM has the potential to improve full-body predictions, including finger articulation.
In contrast, the second and third rows discard the finger pose from the whole-body model and instead adopt the finger pose directly from the hand-only pose estimator.
Although the hand-only pose estimator remains completely independent of CHAM and unchanged during training, simply replacing the finger pose (second row) already reduces errors across all datasets, demonstrating the strong articulation capability of the hand-only model.
The third row further improves accuracy by also transferring hand shape, addressing the limited hand shape expressiveness of SMPL-X, which jointly encodes body, hands, and face in a shared latent space, while the hand-only model focuses solely on hand shape representation.

\subsection{Comparison to state-of-the-art methods}

Tab.~\ref{table:compare_sota_wb} shows that our Hand4Whole++ outperforms existing whole-body pose estimation methods, particularly in terms of hand accuracy.
Fig.~\ref{fig:qualitative_comparison} and ~\ref{fig:qualitative_comparison_2} show that ours produce much better hands and upper-body kinematics than previous works~\cite{baradel2024multi,cai2023smpler} from challenging in-the-wild images and images with complicated two-hands interactions.
In particular, Multi-HMR~\cite{baradel2024multi} produces unstable results when the upper body is truncated.
Tab.~\ref{table:compare_sota_hand} further demonstrates that Hand4Whole++ achieves competitive or superior performance compared to state-of-the-art hand-only models, even on challenging benchmarks involving interacting hands.
Notably, although IntagHand~\cite{li2022interacting} and InterWild~\cite{moon2023bringing} are specifically designed for 3D interacting hand reconstruction, Hand4Whole++ achieves comparable MPVPE and significantly lower MRRPE.
The strong MPVPE indicates accurate wrist and finger articulation on par with hand-only methods.
Also, the much lower MRRPE highlights that our upper-body kinematics enable more accurate relative hand positions, as shown in the first row of Fig.~\ref{fig:intro_compare}.
These improvements are largely attributed to CHAM, which refines wrist orientation and overall upper-body structure within the full-body context.
In contrast, single-hand reconstruction methods~\cite{pavlakos2024reconstructing,potamias2024wilor} suffer from huge MRRPE as they do not have full-body context, as shown in the first row of Fig.~\ref{fig:intro_compare}.
Most importantly, all hand pose estimators in Tab.~\ref{table:compare_sota_hand} do not have full-body results, limiting their applicability in broader scenarios.

All results are obtained by running the official released code and weights of each method.
For Multi-HMR~\cite{baradel2024multi}, we utilized GT bounding boxes as its built-in detector failed to localize humans in half of the cases shown in Fig.~\ref{fig:qualitative_comparison} and~\ref{fig:qualitative_comparison_2}.
Note that some previous works~\cite{lin2023one,cai2023smpler} report slightly different numbers on AGORA and ARCTIC in their papers as they use customized splits and neutral-gender SMPL-X fits to obtain evaluation targets, whereas we use GT shape and gender for evaluation targets.
Moreover, some whole-body methods~\cite{moon2022accurate} report hand benchmark performance~\cite{Freihand2019} using separately trained hand-only models, rather than hand outputs from the whole-body mesh.
In contrast, for all methods including ours, we directly evaluate the hand vertices produced by full-body model, thus providing a more faithful assessment of hand accuracy from whole-body predictions.
HMR-Adapter~\cite{shen2024hmr} could not be evaluated as its code is not publicly available, but we expect it to be unstable on interacting hand images due to its disregard for relative hand positions.

%% file: src/conclusion.tex
\section{Conclusion}

\noindent\textbf{Limitations.}
Due to the absence of full-body annotations in hand-only datasets~\cite{moon2020interhand2,moon2023dataset}, non-hand joints are weakly supervised and may be misaligned with the input image, even when the hand predictions are accurate.
Also, Hand4Whole++ relies on two pre-trained models—one for hand pose estimation and another for whole-body pose estimation—which results in increased runtime.

\noindent\textbf{Summary.}
We presented Hand4Whole++, a modular framework that combines pre-trained whole-body and hand pose estimators for accurate 3D whole-body pose estimation. CHAM improves wrist orientation by modulating whole-body features with hand-specific cues, while finger articulation and shape are transferred via rigid alignment. This decoupled design leverages each estimator’s strength without retraining. Beyond standard benchmarks, our preliminary observations on hand-centric images, such as egocentric views, suggest promising practical utility. We plan to further investigate and formally validate these scenarios as future work. Experiments show that Hand4Whole++ significantly boosts hand accuracy and overall pose quality.

%% file: main_suppl.tex
\begin{center}
\textbf{\large Supplementary Material for \\ \vspace{2mm}
\large{``Enhancing Hands in 3D Whole-Body Pose Estimation with \\Conditional Hands Modulator''}}
\end{center}

\setcounter{section}{0}
\setcounter{table}{0}
\setcounter{figure}{0}
\renewcommand{\thesection}{S\arabic{section}}   
\renewcommand{\thetable}{S\arabic{table}}   
\renewcommand{\thefigure}{S\arabic{figure}}

In this supplementary material, we provide more experiments, discussions, and other details that could not be included in the main text due to the lack of pages.
The contents are summarized below:
\begin{itemize}[nosep]
    \item Sec.~\ref{sec:ablation_studies_suppl}: Ablation studies 
    \item Sec.~\ref{sec:implementation_details_suppl}: Implementation details 
    \item Sec.~\ref{sec:running_time_analysis_suppl}: Running time analysis 
\end{itemize}

\input{src_suppl/ablation_studies}

\input{src_suppl/implementation_details}

\input{src_suppl/running_time_analysis}

%% file: src_suppl/ablation_studies.tex
\section{Ablation studies}~\label{sec:ablation_studies_suppl}

\begin{figure}[t]
\begin{center}
\includegraphics[width=\linewidth]{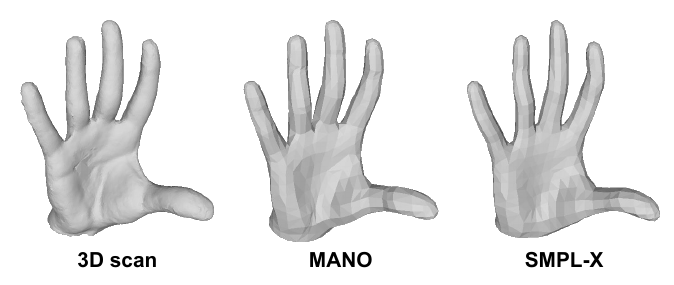}
\end{center}
\vspace*{-5mm}
\caption{
Comparison of hand shape expressiveness between MANO and SMPL-X.
MANO produces hand shapes that more closely match the 3D scans compared to SMPL-X.
}
\label{fig:ablation_shape_transfer_suppl}
\end{figure}

\noindent\textbf{Expressiveness of MANO and SMPL-X shape space.}
In Sec.~\textcolor{cvprblue}{3.3} of the main manuscript, we describe our finger articulation and shape transfer strategy.
The shape transfer aims to leverage the more expressive hand shape space of MANO.
Fig.~\ref{fig:ablation_shape_transfer_suppl} and Tab.~\ref{table:ablation_shape_transfer_suppl} support this design choice by demonstrating that MANO provides a more expressive hand shape representation.
The figure shows that MANO hands better align with the 3D scans, while the table reports lower point-to-point errors compared to SMPL-X hands.

For this comparison, we fit both MANO and SMPL-X models to the 3D scans and keypoints from the MANO test set~\cite{romero2017embodied}.
Both models are initialized by rigidly aligning the wrist and the four MCP joints (index, middle, ring, pinky), with zero shape parameters and zero finger poses.
During SMPL-X optimization, we use only the hand vertices from the full-body mesh.
The objective consists of: 1) 3D keypoint loss, 2) point-to-point loss, and 3) shape parameter regularization, weighted by 1, 1, and 0.001, respectively.
Each sample is optimized for 500 iterations.

\begin{figure}[t]
\begin{center}
\includegraphics[width=\linewidth]{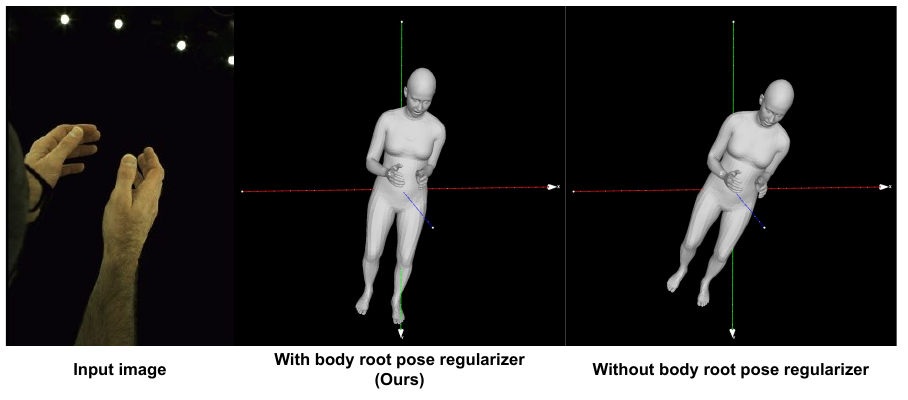}
\end{center}
\vspace*{-5mm}
\caption{
Effectiveness of the body root pose regularizer.
The red, green, and blue axes represent the $x$-, $y$-, and $z$-axes of the world coordinate system, respectively.
The green axis ($y$-axis) points downward, which is opposite to the up-right direction in the world coordinate system.
}
\label{fig:ablation_root_pose_reg_suppl}
\end{figure}

\noindent\textbf{Body root pose regularizer.}
In Sec.~\textcolor{cvprblue}{3.4} of the main manuscript, we describe the body root pose regularizer, which encourages the root pose to remain vertical when only hand annotations are available and full-body annotations are not available.
As shown in Fig.~\ref{fig:ablation_root_pose_reg_suppl}, without this regularizer, the recovered 3D human often exhibits an incorrect root pose due to the lack of full-body annotations in hand-only datasets.
Since hand-only datasets are typically captured with subjects in an upright standing pose, we introduce this prior to regularize the root orientation.
Specifically, we enforce alignment between the up-right direction of the world coordinate system and that of the human body by constraining their dot product to be 1.

\noindent\textbf{Cross-attention in CHAM.}
As described in Sec.~\textcolor{cvprblue}{3.2} of the main manuscript, we introduce cross-attention with 2D positional encoding to capture the inter-relationship between the two hands.
Tab.~\ref{table:ablation_cross_attention_suppl} demonstrates that our cross-attention effectively reduces both MPVPE and MRRPE in hand-only datasets with challenging two-hand interactions.

\begin{table}[t]
\footnotesize
\centering
\setlength\tabcolsep{1.0pt}
\def\arraystretch{1.1}
\caption{Comparison of point-to-point distances between 3D scans and optimized hand meshes.
We report the mean point-to-point distances between 3D scans and hand meshes optimized from 1) SMPL-X and 2) MANO.}
\vspace*{-3mm}
\scalebox{1.0}{
\begin{tabular}{C{4.0cm}|C{4.0cm}}
\specialrule{.1em}{.05em}{.05em}
Settings & Point-to-point error (mm) \\ \hline
SMPL-X hands & 1.98 \\
\textbf{MANO hands (Ours)} & 1.34 \\
 \specialrule{.1em}{.05em}{.05em}
\end{tabular}
}
\label{table:ablation_shape_transfer_suppl}
\end{table}

\begin{table*}[t]
\footnotesize
\centering
\setlength\tabcolsep{1.0pt}
\def\arraystretch{1.1}
\caption{Comparison of MPVPE/MRRPE with and without cross-attention in CHAM.}
\vspace*{-3mm}
\scalebox{1.0}{
\begin{tabular}{C{4.0cm}|C{2.0cm}C{2.0cm}C{2.0cm}}
\specialrule{.1em}{.05em}{.05em}
Settings & IH26M & ReIH & HIC \\ \hline
Without cross attention & 9.77/35.36 & 9.12/19.42 & 18.25/30.44 \\
\textbf{With cross attention (Ours)} &  \textbf{9.40}/\textbf{32.30} & \textbf{7.98}/\textbf{16.37} & \textbf{17.72}/\textbf{29.09} \\
 \specialrule{.1em}{.05em}{.05em}
\end{tabular}
}
\label{table:ablation_cross_attention_suppl}
\end{table*}

%% file: src_suppl/implementation_details.tex
\begin{table*}[t]
\footnotesize
\centering
\setlength\tabcolsep{1.0pt}
\def\arraystretch{1.1}
\caption{Running time (in seconds) to process a single image on an RTX A6000 GPU.}
\vspace*{-3mm}
\scalebox{1.0}{
\begin{tabular}{C{2.0cm}C{4.0cm}C{2.0cm}C{4.5cm}|C{2.0cm}}
\specialrule{.1em}{.05em}{.05em}
Hand detector & Hand pose estimator~\cite{potamias2024wilor} & CHAM & Whole-body pose estimator~\cite{cai2023smpler} & Total \\ \hline
0.01 & 0.05 & 0.01 & 0.03 & 0.1 \\
 \specialrule{.1em}{.05em}{.05em}
\end{tabular}
}
\label{table:running_time_suppl}
\end{table*}

\section{Implementation details}~\label{sec:implementation_details_suppl}

We implement our method using PyTorch. The model is trained for 4 epochs with a batch size of 32, using the Adam optimizer and an initial learning rate of 1e-4. The learning rate is reduced by a factor of 10 at the 3rd epoch. All loss terms are equally weighted with a weight of 1. Training is performed on a single NVIDIA RTX A6000 GPU and takes approximately 20 hours.
All other implementation details are provided in the released code.

%% file: src_suppl/running_time_analysis.tex
\section{Running time analysis}~\label{sec:running_time_analysis_suppl}

Tab.~\ref{table:running_time_suppl} summarizes the runtime of each component in our Hand4Whole++.
Overall, Hand4Whole++ runs at 10 frames per second on a single RTX A6000 GPU.
Among all components, the hand pose estimator (WiLoR\cite{potamias2024wilor}) is the most time-consuming.
In contrast, our CHAM module runs significantly faster than both the hand pose estimator (WiLoR~\cite{potamias2024wilor}) and the whole-body pose estimator (SMPLer-X~\cite{cai2023smpler}), thanks to its lightweight design.